\documentclass[10pt,twocolumn,letterpaper]{article}

\usepackage{iccv}
\usepackage{times}
\usepackage{epsfig}
\usepackage{graphicx}
\usepackage{amsmath}
\usepackage{amssymb}
\usepackage{amsmath}
\usepackage{amssymb}
\usepackage{algorithm}
\usepackage{algorithmicx}
\usepackage{algpseudocode}
\usepackage{graphics}
\usepackage{threeparttable}
\usepackage{color}
\usepackage[normalem]{ulem}
\usepackage{multirow}
\usepackage{float}
\usepackage{amsfonts}
\usepackage{bm}
\usepackage{array}
\usepackage[table]{xcolor}
\usepackage{authblk}

\usepackage[breaklinks=true,bookmarks=false]{hyperref}

\iccvfinalcopy 

\begin{document}

\title{Super-Trajectory for Video Segmentation}
\author[1]{Wenguan Wang~}
\author[1]{Jianbing Shen\thanks{Corresponding author: Jianbing Shen (shenjianbing@bit.edu.cn). This work was supported in part by the National Basic Research Program of China (973 Program) (No. 2013CB328805), the National Natural Science Foundation of China (61272359),  the Australian Research Council's Discovery Projects funding scheme (project DP150104645), and the Fok Ying-Tong Education Foundation for Young Teachers. Specialized Fund for Joint Building Program of Beijing Municipal Education Commission.}~}
\author[2]{Jianwen Xie~}
\author[3]{Fatih Porikli~}
\affil[1]{\small Beijing Lab of Intelligent Information Technology, School of Computer Science, Beijing Institute of Technology, China}
\affil[2]{\small University of California, Los Angeles, USA}
\affil[3]{\small Research School of Engineering, Australian National University, Australia}

\maketitle
\thispagestyle{empty}
\pagestyle{empty}

\begin{abstract}
We introduce a novel semi-supervised video segmentation approach based on an efficient video representation, called as ``super-trajectory''. Each super-trajectory corresponds to a group of compact trajectories that exhibit consistent motion patterns, similar appearance and close spatiotemporal relationships.
We generate trajectories using a probabilistic model, which handles occlusions and drifts in a robust and natural way. To reliably group trajectories, we adopt a modified version of the density peaks based clustering algorithm that allows capturing rich spatiotemporal relations among trajectories in the clustering process. The presented video representation is discriminative enough to accurately propagate the initial annotations in the first frame onto the remaining video frames. Extensive experimental analysis on challenging benchmarks demonstrate our method is capable of distinguishing the target objects from complex backgrounds and even reidentifying them after occlusions. 
\end{abstract}

\section{Introduction}
\label{sec:introduction}
\begin{figure}
 \label{fig:fig1} 
  \centering
      \includegraphics[width=0.98 \linewidth]{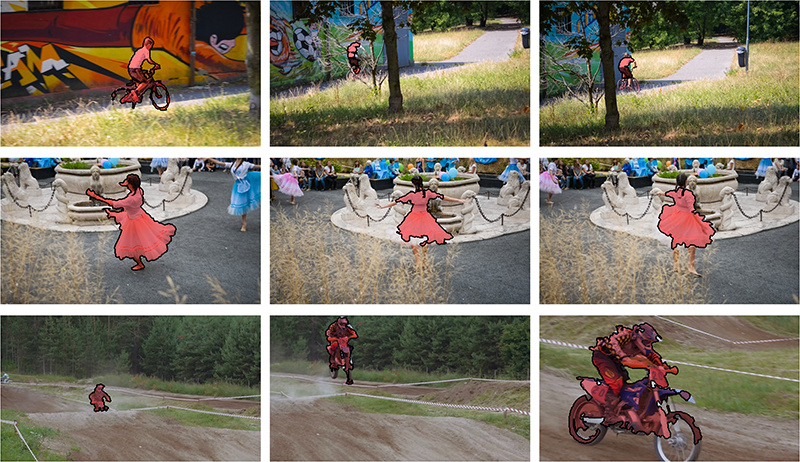}
\caption{Our video segmentation method takes the first frame annotation as initialization (left). Leveraging on  super-trajectories, the segmentation process achieves superior results even for challenging scenarios including heavy occlusions, complex appearance variations, and large shape deformations (middle, right). }
\end{figure}

We state the problem of semi-supervised video object segmentation as the partitioning of objects in a given video sequence with available annotations in the first frame. Aiming for this task, we incorporate an efficient video representation, \textit{super-trajectory}, to capture the underlying spatiotemporal structure information that is intrinsic to real-word scenes. Each super-trajectory corresponds to a group of trajectories that are similar in nature and have common characteristics. A point trajectory, \textit{e.g.}, the tracked positions of an individual point across multiple frames, is a constituent of super-trajectory. This representation captures several aspects of a video:
\vspace{-2mm}
\begin{itemize}
  \item \textit{Long-term motion information} is explicitly modeled as it consists of trajectories over extended periods;
  \vspace{-2mm}
  \item \textit{Spatiotemporal location information} is implicitly interpreted by clustering nearby trajectories; and
  \vspace{-2mm}
  \item \textit{Compact features}, such as color and motion pattern, are described in a conveniently compact form.
\end{itemize}
With above good properties, super-trajectory simplifies and reduces the complexity of propagating human-provided labels in the segmentation process. We first generate trajectory based on Markovian process, which handles occlusions and drifts naturally and efficiently. Then a density peaks based clustering (DPC) algorithm \cite{rodriguez2014} is modified for obtaining reasonable division of the trajectories, which offers proper split of videos in space and time axes.
The design of our super-trajectory is motivated by the flowing two aspects.

Firstly, for the task of video segmentation, it is desirable to have a powerful abstraction of videos that is robust to structure variations and deformations in image space and time.
As demonstrated in recently released DAVIS dataset~\cite{Perazzi2016}, most of the existing approaches exhibit severe limitations for occlusions, motion blur, and appearance changes. The proposed super-trajectory, encoded with several well properties, is able to capture above instances (see Fig.~\ref{fig:fig1}).

Secondly, from the perspective of feature generation, point trajectory is desired to be improved for meeting above requests.
Merging and splitting video segments (and corresponding trajectories) into atomic spatiotemporal components is essential for handling occlusions and temporal discontinuities. However, it is well-known that,  classical clustering methods (\textit{e.g.}, k-means and spectral clustering), which are widely adopted by previous trajectory methods, even cannot reached a consensus on the definition of a cluster.
Here, we modify DPC algorithm \cite{rodriguez2014} for grouping trajectories, favoring its advantage of choosing cluster centers based on a reasonable criterion.

We conduct video segmentation via operating trajectories as unified super-trajectory groups. To eliminate adverse effects of camera motion, we introduce a \textit{reverse-tracking} strategy to exclude objects that originate outside the frame. To reidentify objects after occlusions, we exploit \textit{object re-occurrence} information, which reflects the spatiotemporal relations of objects across the entire video sequence.

The remainder of the article is organized as follows.
A summarization of related work is introduced in Sec.~\ref{sec:relatedwork}.
Our approach for super-trajectory generation is presented in detail in Sec.~\ref{sec:supertrajectory}. In Sec.~\ref{sec:segmentation}, we describe our super-trajectory based video segmentation algorithm.
We then experimentally demonstrate its robustness, effectiveness, and efficiency in Sec.~\ref{sec:Experimentalresults}.
Finally, we draw conclusions in Sec.~\ref{sec:conclusion}.

\section{Related Work}
\label{sec:relatedwork}
We provide a brief overview of recent works in video object segmentation and point trajectory extraction.
\subsection{Video Object Segmentation}
According to the level of supervision required, video segmentation techniques can be broadly categorized as unsupervised, semi-supervised and supervised methods.

Unsupervised algorithms \cite{vazquez2010multiple,xu2012streaming,oneata2014spatio, wang2015consistent,wang2015robust} do not require manual annotations but often rely on certain limiting assumptions about the application scenario. Some techniques \cite{Brox2010,Fragkiadaki12,Papazoglou2013} emphasize the importance of motion information. More specially, \cite{Brox2010,Fragkiadaki12} analyze long-term motion information via trajectories, then solve the segmentation as a trajectory clustering problem. The works \cite{faktor2014,Wang2015,Wang2017} introduce saliency information \cite{wang2016correspondence} as prior knowledge to infer the object.  Recently, \cite{Lee2011,ma2012,zhang2013,fragkiadaki2015learning,Xiao_2016_CVPR} generate object segments via ranking several object candidates.

Semi-supervised video segmentation, which also refers to \textit{label propagation}, is usually achieved via propagating
human annotation specified on one or a few key-frames onto the entire video sequence \cite{brendel2009,Badrinarayanan2010,tsai2010,budvytis2011,li2013,hariharan2015hypercolumns,shankar2015,Tsai2016}.
These methods mainly use flow-based random field propagation models \cite{Sudheendra2012}, patch-seams based propagation strategies \cite{Ramakanth2014}, energy optimizations over graph models \cite{Perazzi2015}, joint segmentation and detection frameworks \cite{Wen_2015_CVPR}, or pixel segmentation on bilateral space \cite{maerki2016}.

Supervised methods \cite{Bai2009,zhong2012discontinuity,Fan2015} require tedious user interaction and iterative human corrections. These methods can attain high-quality boundaries while suffering from extensive and time-consuming human supervision.
\subsection{Point Trajectory}
Point trajectories are generated through tracking points over multiple frames and have the advantage of representing long-term motion information. Kanade-Lucas-Tomasi (KLT) \cite{shi1994good} is among the most popular methods that track a small amount of feature points. Inspiring several follow-up studies in video segmentation and action recognition, optical flow based dense trajectories \cite{sundaram2010dense} improve over sparse interest point tracking.
In particular, \cite{wang2011,wang2013,wang2015action} introduce dense trajectories for action recognition. Other methods
\cite{Brox2010,fragkiadaki2011detection,lezama2011track,fragkiadaki2012two,Fragkiadaki12,ochs2012higher,ochs2014segmentation,keuper2015motion,chen2015video} address the problem of unsupervised video segmentation, in which
case the problem also be described as \textit{motion segmentation}. These methods usually track points via dense optical flow and perform segmentation via clustering trajectories.

Existing approaches often handle trajectories in pairs or individually and directly group all the trajectories into few clusters as segments, easily ignoring the inner coherence in a group of similar trajectories. Instead, we operate trajectories as united super-trajectory groups instead of individual entities, thus offering compact and atomic video representation and fully exploiting spatiotemporal relations among trajectories.

\section{Super-Trajectory via Grouping Trajectories}
\label{sec:supertrajectory}
\subsection{Trajectory Generation}
\label{sec:trajectorygeneration}
Given a sequence of video frames $I_{1:T} = \{I_1, {\cdots},I_T\}$ within time range $[1,T]$, each pixel point can be tracked to the next frame using optical flow. This tracking process can be executed frame-by-frame until some termination conditions (\textit{e.g.}, occlusion, incorrect motion estimates, \textit{etc.}) are reached. The tracked points are composed into a trajectory and a new tracker is initialized where prior tracker finished. We build our trajectory generation on a unified probabilistic model which naturally considers various termination conditions.

Let $\textbf{w}$ denote a flow field indexed by pixel positions that returns a 2D flow vector
at a given point. Using LDOF~\cite{BroxT2011}, we compute forward-flow field $\textbf{w}_t$ from frame $I_t$ to $I_{t+1}$, and
the backward-flow field $\hat{\textbf{w}}_t$ from $I_t$ to $I_{t-1}$. We track pixel potion $\textbf{x} = (x,y,t)$
to the consecutive frames in both directions. The tracked points of consecutive frames are concatenated to form a trajectory $\tau$:
 \begin{equation}
    \begin{aligned}
    \tau = \{\textbf{x}_n\}_{n=1}^L= \{(x_n,y_n,t_n)\}_{n=1}^L,  ~~~~~~t_n \in [1,T],
    \end{aligned}
    \label{eq:1}
\end{equation}
where $L$ indicates the length of trajectory $\tau$ and $(x_n,y_n)=(x_{n-1},y_{n-1})\!+\!\textbf{w}_{t_{n\!-\!1}}(x_{n-1},y_{n-1})$.
We model point tracking process as a first order Markovian process, and denote the probability that $n$-th point $\textbf{x}_n$ of trajectory $\tau$ is correctly tracked from frame $I_{t_1}$ as $p(\textbf{x}_n|I_{t_1:t_n})$. The prediction model is defined by:
 \begin{equation}
    \begin{aligned}
    p(\textbf{x}_n|I_{t_1:t_n}) = p(\textbf{x}_n|\textbf{x}_{n-1})p(\textbf{x}_{n-1}|I_{t_1:t_{n-1}}),
    \end{aligned}
    \label{eq:2}
\end{equation}
where $p(\textbf{x}_{1}|I_{t_1}) = 1$ and $p(\textbf{x}_n|\textbf{x}_{n-1})$ is formulated as:
 \begin{equation}
    \begin{aligned}
    p(\textbf{x}_n|\textbf{x}_{n-1}) = exp\{-(\mathcal{E}_{app}+\mathcal{E}_{occ})\}.
    \end{aligned}
    \label{eq:3}
\end{equation}
The energy functions $\mathcal{E}$ penalize various potential tracking error. The former energy $\mathcal{E}_{app}$ is expressed as:
 \begin{equation}
    \begin{aligned}
    \mathcal{E}_{app}(\textbf{x}_n,\textbf{x}_{n\!-\!1}) = ||I_{t_n}(x_n,y_n) - I_{t_{n\!-\!1}}(x_{n\!-\!1},y_{n\!-\!1})||,
    \end{aligned}
    \label{eq:4}
\end{equation}
which penalizes the appearance variations between corresponding points. The latter energy $\mathcal{E}_{occ}$ is included to penalize occlusions. It uses the consistency of the forward and backward flows:
\begin{small}
 \begin{equation}
    \begin{aligned}
    \mathcal{E}_{occ}(\textbf{x}_n,\textbf{x}_{n\!-\!1})\! =\! \frac{|| \hat{\textbf{w}}_{t_n}(x_n,y_n)\!+\! \textbf{w}_{t_{n\!-\!1}}(x_{n\!-\!1},y_{n\!-\!1}) ||}{||\hat{\textbf{w}}_{t_n}(x_n,y_n)|| \!+\! ||\textbf{w}_{t_{n\!-\!1}}(x_{n\!-\!1},y_{n\!-\!1}) ||     }.
    \end{aligned}
    \label{eq:5}
\end{equation}
\end{small}
When this consistency constraint is violated, occlusions or unreliable optical flow estimates might occur (see \cite{sundaram2010dense} for more discussion). It is important to notice that the proposed tracking model performs accurately yet our model is not limited to the above constraints. We terminate the tracking process when $p(\textbf{x}_n|I_{t_1:t_n}) < 0.5$, and then we start a new tracker at $\textbf{x}_n$. In our implementation, we discard the trajectories shorter than four frames.
\subsection{Super-Trajectory Generation}
\label{sec:supertrajectorygeneration}
Previous studies indicate the value of trajectory based representation for long-term motion information. Our additional intuition is that neighbouring trajectories exhibit compact spatiotemporal relationships and they have similar natures in appearance and motion patterns. This motives us operating on trajectories as united groups.

We generate super-trajectory by clustering trajectories with density peaks based clustering (DPC) algorithm\cite{rodriguez2014}. Before introducing our super-trajectory generation method, we first describe DPC.

\noindent\textbf{Density Peaks based Clustering (DPC)~}
DPC is proposed to cluster the data by finding of density peaks. It provides a unique solution of fast clustering based on the idea that cluster centers are characterized by a higher density than their neighbors and by a relatively large distance from points with
higher densities. It offers a reasonable criterion for finding clustering centers.

Given the distances $d_{ij}$ between data points, for each data point $i$, DPC calculates two quantities: local density $\rho_i$ and its distance $\delta_i$ from points of higher density. The local density $\rho_i$ of data point $i$ is defined as \footnote{Here we do not use the cut-off kernel or gaussian kernel adopted in \cite{rodriguez2014}, due to the small data amount.}:
 \begin{equation}
    \begin{aligned}
    \rho_i = \sum\nolimits_{j} d_{ij}.
    \end{aligned}
    \label{eq:6}
\end{equation}
Here, $\delta_i$ is measured by computing the minimum distance between the point $i$ and any other
point with higher density:
 \begin{equation}
    \begin{aligned}
    \delta_i = \min_{j:\rho_j>\rho_i} (d_{ij}).
    \end{aligned}
    \label{eq:7}
\end{equation}
For the point with highest density, it takes $\delta_i = max_j(d_{ij})$. 

Cluster centers are the points with high local density ($\rho \uparrow$) and large distance ($\delta \uparrow$) from other points with higher local density.
The data points can be ranked via $\gamma_i = \rho_i \delta_i$, and the top ranking points are selected as centers.
After successfully declaring cluster centers, each remaining
data points is assigned to the cluster center as its nearest neighbor of higher density.

\noindent\textbf{Grouping Trajectories via DPC~}
Given a trajectory $\tau: \{(x_n,y_n,t_n)\}_n$ spans $L$ frames, we define three features: spatial location ($l_{\tau}$), color ($c_{\tau}$), and velocity ($v_{\tau}$), for describing $\tau$:
 \begin{equation}
    \begin{aligned}
    &l_{\tau} = \frac{1}{L}\sum^L_{n=1}(x_n,y_n), ~~c_{\tau} = \frac{1}{L}\sum^L_{n=1}I_{t_n}(x_n,y_n),\\
    &v_{\tau} = \frac{1}{L}\sum^L_{n=1}\big(\frac{1}{\Delta t}(x_{n+\Delta t}-x_n,y_{n+\Delta t}-y_n)\big),
    \end{aligned}
    \label{eq:8}
\end{equation}
where we set $\Delta t = 3$. We tested $\Delta t = \{5,7,9\}$ and did not observe obvious effect on the results.

Between each pair of trajectories $\tau_i$ and $\tau_j$ that share some frames,
we define their distance $d_{ij}$ via measuring descriptor similarity:
\begin{equation}
    \begin{aligned}
    d_{ij} = \sum\nolimits_{f\in\{l,c,v\}}||f_{\tau_i} - f_{\tau_j}||.
    \end{aligned}
    \label{eq:9}
\end{equation}
We normalize color distance on max intensity, location distance on sampling step $R$ (detailed below), motion distance on the mean motion magnitude of all the trajectories, which makes above distance measures to have similar scales.
In case there is no temporal overlap, we set $d_{ij} = H$, where $H$ has a very large value.

\begin{figure}
  \centering
     \includegraphics[width=0.95 \linewidth]{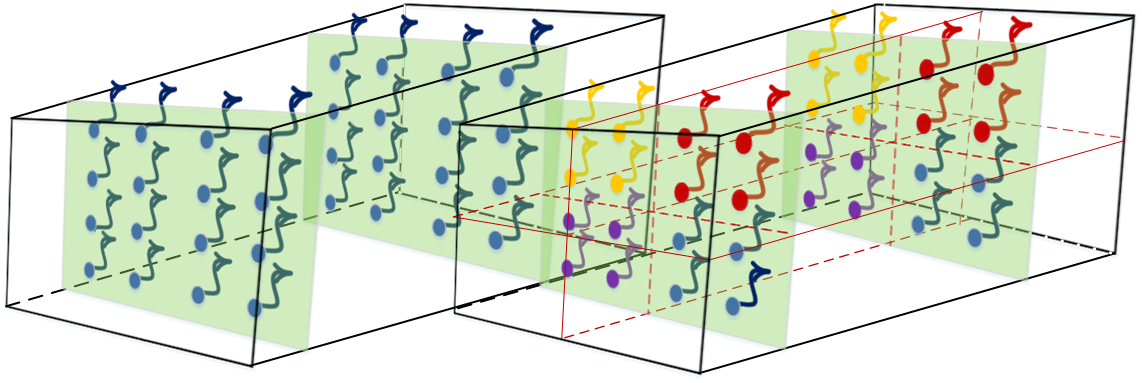}
     \hfill\mbox{} \\
     \mbox{}\hfill (a) \hfill\mbox{}
     \mbox{}\hfill (b) \hfill\mbox{}
\caption{Illustration of initial super-trajectory generation. (a) The arrows indicate trajectories and the dots indicate the initial location of trajectory. (b) We roughly divide all the trajectories into $K$ groups with a given number of spatial grids $K = 4$.}
 \label{fig:fig2} 
\end{figure}

We first roughly partition trajectories into several non-overlap clusters, and then iteratively updates each partition to get the optimized trajectory clusters.

The only parameter of our super-trajectory algorithm is number of spatial grids $K$, as the degree of spatial subdivision. The spatial sampling step becomes $R\! = \!\sqrt{S/K}$, where $S$ refers to the product of the height and width of image frame. The clustering procedure begins with an initialization step where we divide the input video $I_{1:T}$ into several non-overlap spatiotemporal volumes of size $R\times R\times T$. As shown in Fig.~\ref{fig:fig2}, all trajectories $\mathcal{T}\! =\!\{\tau_i\}_i$ are divided into $K$ volumes. A trajectory $\tau$ falls into the volume where it starts.
Then we need to find a proper cluster number of each trajectory group, thereby further offering a reasonable temporal split of video.

For each trajectory group, we initially estimate the cluster number as $C = T/\overline{L}$, where $\overline{L}$ indicates
the average length of all the trajectories. Then we apply a modified DPC algorithm for generating trajectory clusters, as described with in Alg.~\ref{alg:alg1}. In Alg.~\ref{alg:alg1}-3, if we have $\delta_i = H$, then trajectory $\tau_i$ does not have any temporal overlap with those trajectories have higher local densities. That means trajectory $\tau_i$ is the center of a isolated group.  If $C<n'$, in Alg.~\ref{alg:alg1}-4,
\begin{algorithm}
\caption{DPC for Generating Super-Trajectory Centers}
\label{alg:alg1}
\begin{algorithmic}[1]
\Require A sub-group of trajectories $\mathcal{T'}\!=\!\{\tau'_i\}_{i}$ ($\mathcal{T'} \subset \mathcal{T}$), distance matrix $\{d_{ij}\}$ via Eq.~\ref{eq:9} and cluster number $C$;
\Ensure Organized trajectory clusters;
\State Compute local densities $\{\rho_i\}_{i}$ via Eq.~\ref{eq:6};
\State Compute distance $\{\delta_i\}_{i}$ via Eq.~\ref{eq:7};
\State Find $\{\tau'_{i'}\}_{i'}$ with $\delta_{i'} = H$, where $|\{\tau'_{i'}\}_{i'}|=n'$;
\If  {$C<n'$}
\State Select $\{\tau'_{i'}\}_{i'}$ as cluster centers;
\Else
\State Compute $\{\gamma_i\}_{i}$ via $\gamma_i = \rho_i \delta_i$;
\State Select the trajectories with $C$ highest $\gamma$ values as $~~~~~~~$cluster centers;
\EndIf
\State Assign remaining trajectories to cluster centers.
\end{algorithmic}
\end{algorithm}
that means there exist more than $C$ unconnected trajectory groups. Then we select the trajectories with highest densities of those unconnected trajectory groups as centers (Alg.~\ref{alg:alg1}-5).
Otherwise, in Alg.~\ref{alg:alg1}-7,8, the trajectories with the $C$ highest $\gamma$ values are selected as the cluster centers.
The whole initialization process is described in Alg.~\ref{alg:alg2}-1,2,3.

\begin{figure}
  \centering
     \includegraphics[width=0.98 \linewidth]{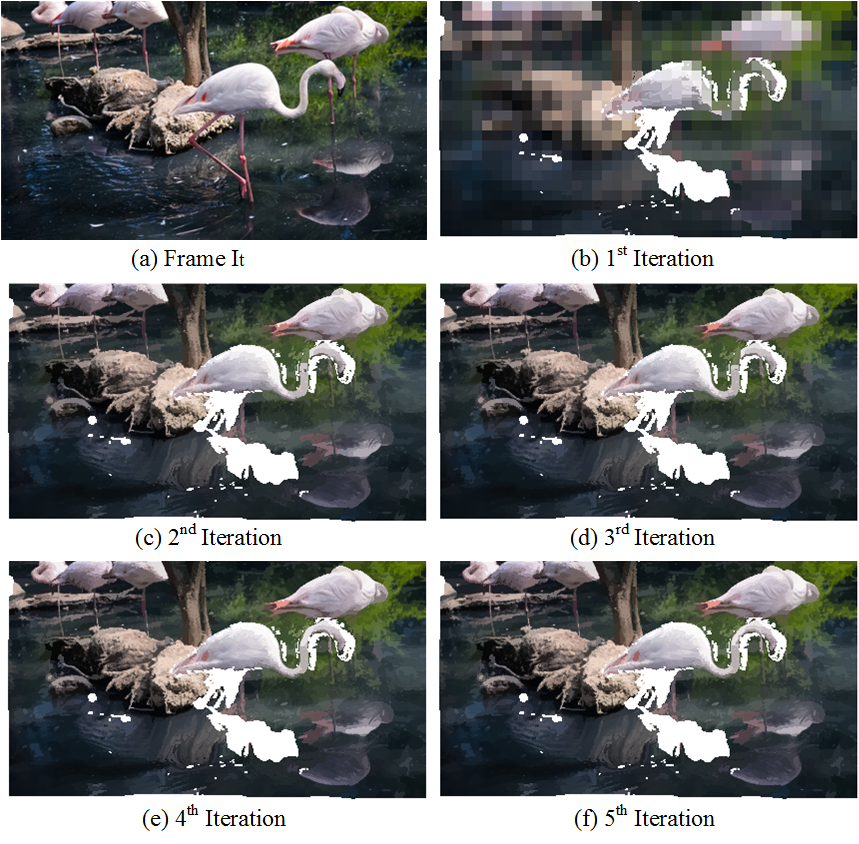}
\caption{Illustration of our super-trajectory generation via iterative trajectory clustering. (a) Frame $I_t$. (b)-(f) Visualization results of super-trajectory in time slice $I_t$ with different iterations. Each pixel is assigned the average color of all the points along the trajectory which it belongs to. The blank areas are the discarded trajectories which are shorter than four frames.}
\label{fig:fig3} 
\end{figure}

Based on the above initialization process, we group trajectories into super-trajectories according to their spatiotemporal relationships and similarities (see Fig.~\ref{fig:fig3}(b)). Next, we iteratively refine our super-trajectory assignments. In this process, each trajectory is classified into the nearest cluster center. For reducing the searching space, we only search the trajectories fall into a $2R \times 2R \times T$ space-time volume around the cluster center $\tau_{i'}$ (Alg.~\ref{alg:alg2}-7).
This results in a significant speed advantage by limiting the size of search space to reduce the number of distance calculations.
Once each trajectory has been associated to the nearest cluster center, an update step adjusts the center of each trajectory cluster via Alg.~\ref{alg:alg1} with $C = 1$ (Alg.~\ref{alg:alg2}-14,15). We drop very small trajectory clusters and combine those trajectories to other nearest trajectory clusters.
In practice, we find 5 iterations for above refining process are enough for obtaining satisfactory performance. Visualization results of super-trajectory generation with different iterations are presented in Fig.~\ref{fig:fig3}.

Using Alg.~\ref{alg:alg1}, we group all trajectories $\mathcal{T}\!=\!\{\tau_i\}_{i}$ into $m$ nonoverlap clusters, represented as super-trajectories $\mathcal{X}\!=\!\{\chi_j\}_{j=1}^{m}$, where $\chi_j = \{\tau_i~|~\tau_i \text{~is classified into~} j \text{-th cluster}\\ \text{via Alg.~\ref{alg:alg2}}\}$. It is worth to note that, $m$ (the number of super-trajectories) is varied in each iteration in Alg.~\ref{alg:alg2} since we merge small clusters into other clusters. Additionally, $m$ for different videos is different even with same input parameter $K$. That is important, since different videos have different temporal characteristics, thus we only constrain their spatial shape via $K$.
\begin{algorithm}
\caption{Super-Trajectory Generation}
\label{alg:alg2}
\begin{algorithmic}[1]
\Require All the trajectories $\{\tau_i\}_{i}$, spatial sampling step $R$;
\Ensure Super-trajectory assignments; $~~~~~~~~~~~~~~~~~~~~~~~~$
\text{/*~Initialization */}
\State Obtain $K$ trajectory groups via spatial sampling step $R$;
\State Set initial cluster number $C = T/\overline{L}$ for each group;
\State Obtain initial cluster centers $\{\tau_{i'}\}_{i'}$ from each trajectory group via Alg.~\ref{alg:alg1}, where $|\{\tau_{i'}\}_{i'}|=m$;
\Loop{$~~~~~~~~~~~~~~~~~~~~~~~~~~~~~~~~~~~~~~~~~~~~~~~~~~~~~~~~~~$}
\text{~~~~~~/*~Iterative Assignment */}
\State Set label $l_i = -1$ and distance $\kappa_i = H$ for each $~~~~~~~$trajectory $\tau_i$;
\For  {each trajectory cluster center $\tau_{i'}$}
\For  {each trajectory \!$\tau_j$ \!falls \!in \!a \!$2R \!\times\! 2R \!\times\! T$ $~~~~~~~~~~~~~$\!space-time volume around $\tau_{i'}$}
\State Compute distance $d_{ji'}$ between $\tau_j$ and $\tau_{i'}$ $~~~~~~~~~~~~~~~~~~$ via Eq.~\ref{eq:9};
\If {$d_{ji'}<\kappa_j$}
\State Set $\kappa_j =d_{ji'} $, $l_j = i'$;
\EndIf
\EndFor
\EndFor {$~~~~~~~~~~~~~~~~~~~~~~~~~~~~~~~~~~~~~~~~~~~~~~~~~~~~~~~~~~$}
\text{~~~~~~/*~Update Assignment */}
\State Set cluster number $C = 1$ for each group;
\State Update $\{\tau_{i'}\}_{i'}$ for each cluster via Alg.~\ref{alg:alg1}.
\EndLoop
\end{algorithmic}
\end{algorithm}
\vspace{-3mm}
\section{Super-Trajectory based Video Segmentation}
\label{sec:segmentation}
In Sec.~\ref{sec:supertrajectory}, we cluster a set of compact trajectories into super-trajectory. In this section, we describe our video segmentation approach that leverages on super-trajectories.

Given the mask $\mathcal{M}$ of the first frame, we seek a binary
partitioning of pixels into foreground and background classes. Clearly, the annotation can be
propagated to the rest of the video, using the trajectories that start at the first frame. However, only a few of points can be successfully tracked across the whole scene, due to occlusion, drift or unreliable motion estimation. Benefiting from our efficient trajectory clustering approach, super-trajectories are able to spread more annotation information over longer periods. This inspires us to base our label propagation process on super-trajectory.

\begin{figure*}
  \centering
     \includegraphics[width=0.98 \linewidth]{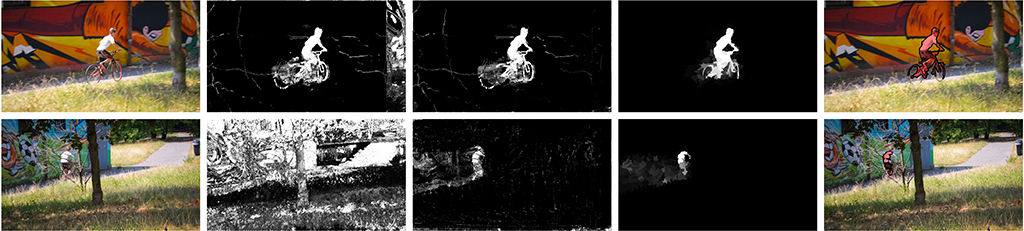}
     \hfill\mbox{} \\
     \mbox{}\hfill (a) \hfill\mbox{}
     \mbox{}\hfill (b) \hfill\mbox{}
     \mbox{}\hfill (c) \hfill\mbox{}
     \mbox{}\hfill (d) \hfill\mbox{}
     \mbox{}\hfill (e) \hfill\mbox{}
\caption{(a) Input frames. (b) Foreground estimates via Eq.~\ref{eq:10}. (c) Foreground estimates via our reverse tracking strategy (Eq.~\ref{eq:12}). (d) Foreground estimates via backward re-occurrence based optimization (Eq.~\ref{eq:14}). (e) Final segmentation results.}
 \label{fig:fig4}
\end{figure*}

For inferring the foreground probability of super-trajectories $\mathcal{X}$, we first divide all the trajectories $\mathcal{T}$ into three categories: foreground trajectories $\mathcal{T}^f$, background trajectories $\mathcal{T}^b$ and unlabeled trajectories $\mathcal{T}^u$, where $\mathcal{T} = \mathcal{T}^f\cup\mathcal{T}^b\cup\mathcal{T}^u$. The $\mathcal{T}^f$ and $\mathcal{T}^b$ are the trajectories which start at the first frame and are labeled by the annotation mask $\mathcal{M}$, while the $\mathcal{T}^u$ are the trajectories start at any frames except the first frame, thus cannot be labeled via $\mathcal{M}$. Accordingly, super-trajectories $\mathcal{X}$ are classified into two categories: labeled ones $\mathcal{X}^l$ and unlabeled ones $\mathcal{X}^u$. A labeled super-trajectory $\chi^l_j \in \mathcal{X}^l$ contains at least one labeled trajectory from $\mathcal{T}^f$ or $\mathcal{T}^b$, and its foreground probability can be computed
as the ratio between the included foreground trajectories and the labeled ones it contains:
\begin{equation}
    \begin{aligned}
    p_f(\chi^l_j) = \frac{|\chi^l_j \cap \mathcal{T}^f|}{|\chi^l_j \cap\mathcal{T}^f|+|\chi^l_j \cap\mathcal{T}^b|}.
    \end{aligned}
    \label{eq:10}
\end{equation}
For the points belonging to the labeled super-trajectory $\chi^l_j$, their foreground probabilities are set as $p_f(\chi^l_j)$.

Then we build an appearance model for estimating the foreground probabilities of unlabeled pixels. The appearance model
is built upon the labeled super-trajectories $\mathcal{X}^l$, consists of two weighted Gaussian Mixture Models over RGB colour values, one for the foreground and one for the background.
The foreground GMM is estimated form all labeled super-trajectories $\mathcal{X}^l$, weighted by their foreground probabilities $\{p_f(\chi^l_j)\}_j$.
The estimation of background GMM is analogous, with the weight replaced by the background probabilities $\{1\!-\!p_f(\chi^l_j)\}_j$.
The appearance models leverage the foreground and
background super-trajectories over many frames, instead of
only using the first frame or labeled trajectories, therefore they can robustly estimate
appearance information.

Although above model successfully propagates more annotation information across the whole video sequence, it still suffers from some difficulties: the model will be confused when a new object come into view (see Fig.~\ref{fig:fig4} (b)). To this, we propose to \textit{reverse track points} for excluding new incoming objects. We compute the `source' of unlabeled trajectory $\tau_i^u\in\mathcal{T}^u$:
\begin{equation}
    \begin{aligned}
    (x_0,y_0) = (x_1,y_1)-v_{\tau_i^u},
    \end{aligned}
    \label{eq:11}
\end{equation}
where $(x_1,y_1)$ indicates starting position and $v_{\tau_i^u}$ refers to velocity via Eq.~\ref{eq:8}. It is clear that, if the virtual position $(x_0,y_0)$ is out of image frame domain, trajectory $\tau_i^u$ is a latecomer. For those trajectories $\mathcal{T}^o \subset \mathcal{T}^u$ start outside view, we treat them as background. Labeled super-trajectory $\chi^l_j \in \mathcal{X}^l$ is redefined as the one contains at least one trajectory from $\mathcal{T}^f$, $\mathcal{T}^b$ or $\mathcal{T}^o$, and Eq.~\ref{eq:10} is updated as
\begin{equation}
    \begin{aligned}
    p_f(\chi^l_j) = \frac{|\chi^l_j \cap \mathcal{T}^f|}{|\chi^l_j \cap\mathcal{T}^f|+|\chi^l_j \cap\mathcal{T}^b|+|\chi^l_j \cap\mathcal{T}^o|}.
    \end{aligned}
    \label{eq:12}
\end{equation}
Those outside trajectories $\mathcal{T}^o$ are also adopted for training appearance model in prior step. According to our experiment, this assumption offers about $6\%$ performance improvement. Foreground estimation results via our reverse tracking strategy are presented in Fig.~\ref{fig:fig4} (c).


For re-identifying objects after long-term occlusions and constraining segmentation consistency, we explore \textit{re-occurrence} of objects. As suggested by \cite{faktor2014}, objects, or regions, often re-occur both in space and
in time. Here, we build correspondences among re-occurring regions across distant frames and transport foreground estimates globally. This process is based on super-pixel level, since super-trajectories cannot cover all of pixels.


Let $\{r_i\}_i$ be the superpixel set of input video. For each region, we search its $N$ Nearest Neighbors (NNs) as its re-occurring regions using KD-tree search. For region $r_i$ of frame $I_t$, we only search its NNs in previous frames $\{I_1,{\cdots},I_{t}\}$. Such \textit{backward search} strategy is for biasing the segmentation results of prior frames as the propagation accuracy degrades over time. Following \cite{faktor2014}, each region $r_i$ is represented as a concatenation of several descriptors $f_{r_i}$: RGB and LAB color histograms (6 channels$\times$20 bins), HOG descriptor (9 cells$\times$6 orientation bins) computed over a $15\times15$ patch around superpixel center, and spatial coordinate of superpixel center.
The spatial coordinate is with respect to image center and normalized into $[0, 1]$, which implicitly incorporates spatial consistency in NN-search.

\begin{figure*}
  \centering
      \includegraphics[width=0.95 \linewidth]{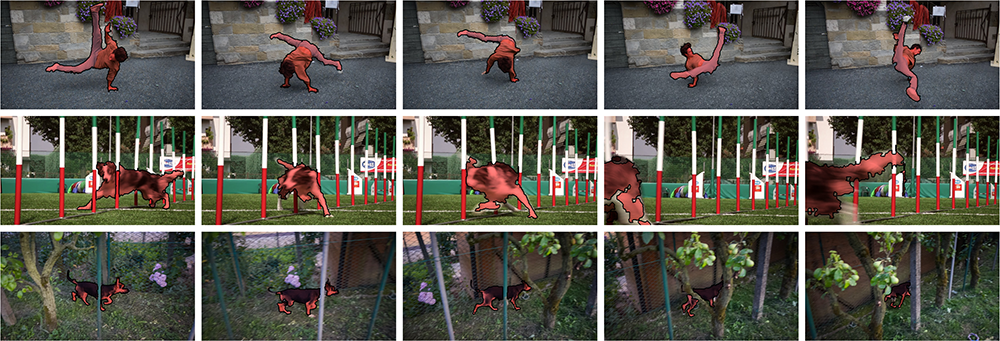}
\caption{Qualitative segmentation results on three video sequences from DAVIS \cite{Perazzi2016} (from top to bottom: \textit{breakdance-flare}, \textit{dog-agility} and \textit{libby}). It can be observed that the proposed algorithm is applicable to a quite general set of sequences and robust to many challenges.}
 \label{fig:fig5} 
  \vspace{-3mm}
\end{figure*}
After NN-search in the feature space, we construct a weight matrix $W$ for all the regions $\{r_i\}_{i}$:
\begin{equation}
    \begin{aligned}
    W_{ij} = \left\{
        \begin{aligned}
            &e^{-||f_{r_i}-f_{r_j}||}~~~~~\text{if } r_j\text{ is one of NNs of }r_i \\
            &1 ~~~~~~~~~~~~~~~~~~~~~~~\text{if } i = j \\
            &0 ~~~~~~~~~~~~~~~~~~~~~~~\text{otherwise }\\
        \end{aligned}
    \right.
    \end{aligned}
    \label{eq:13}
\end{equation}
Then a probability transition matrix $P$ is built via row-wise normalization of $W$. We define a column vector $v$ that gathers all the foreground probabilities of $\{r_i\}_{i}$.  The foreground probability of a superpixel is assigned as the average foreground probabilities of its pixels.

We iteratively update $v$ via the probability transition matrix $P$. In each iteration $k$, we update $v^{(k)}$ via:
\begin{equation}
    \begin{aligned}
    v^{(k)} = Pv^{(k-1)},
    \end{aligned}
    \label{eq:14}
\end{equation}
which equivalents to updating foreground probability of a region with the weighted average of its NNs. 
In each iteration,
we keep the foreground probabilities of those points belonging to labeled trajectories unchanged. Then we recompute $v^{(k)}$ and update it in next iteration. In this way, the relatively accurate annotation information of the labeled trajectories is preserved. Additionally, the annotation information is progressively propagated in a forward way and the super-trajectories based foreground estimates are consistent even across many distant frames (see Fig.~\ref{fig:fig4} (d)).

After 10 iterations, the pixels (regions) with foreground probabilities lager than 0.5 are classified as foreground, thus obtaining final binary segments. In Sec.~\ref{sec:VPA}, we test $N = \{4,6,{\cdots},20\}$ and only observe $\pm0.3\%$ performance variation. We set $N = 8$ for obtaining best performance.
\section{Experimental results}
\label{sec:Experimentalresults}
\noindent\noindent\textbf{Parameter Settings~}
In Sec.~\ref{sec:supertrajectorygeneration},  we set number of spatial grids $K = 1200$. In Sec.~\ref{sec:segmentation}, we over-segment each frame into about $2000$ superpixels via SLIC \cite{Achanta2012} for well boundary adherence. For each superpixel, we set the number of NNs $N=8$. In our experiments, all the parameters of our algorithm are fixed to unity.
\\
\noindent\textbf{Datasets~} We evaluate our method on two public video segmentation benchmarks, namely DAVIS \cite{Perazzi2016}, and Segtrack-V2 \cite{li2013}.
The new released DAVIS \cite{Perazzi2016} contains
50 video sequences (3,~\!\!455 frames in total) and pixel-level manual ground-truth for the foreground object in every frame.
Those videos span a wide range of object segmentation challenges such as occlusions, fast-motion and appearance changes.
Since DAVIS contains diverse scenarios which break classical assumptions, as demonstrated in \cite{Perazzi2016}, most state-of-the-art methods fail to produce reasonable segments.
Segtrack-V2 \cite{li2013} consists of 14 videos with 24 instance objects and 947 frames. Pixel-level mask is offered for every frame.
\subsection{Performance Comparison}\label{sec:PC}

\begin{table}
\centering
\resizebox{0.49\textwidth}{!}{
\setlength\tabcolsep{1.5pt}
\begin{tabular}{|c||c|c|c|c|c|c|c|}  
\hline
\multirow{2}*{Video}
&\multicolumn{7}{c|}{IoU score} \\
\cline{2-8}
&\!\!BVS\!\! &\!\!FCP\!\! &\!\!JMP\!\! &\!\!SEA\!\! &\!\!TSP\!\! &\!\!HVS\!\! &\!\!\!\!\textbf{STV}\!\!\!\!\\
\hline
\hline

breakdance-flare	 &0.727 	&0.723 	&0.430 	&0.131 	&0.040 	&0.499 	&\textbf{0.835} 	\\
camel	             &0.669 	&0.734 	&0.640 	&0.649 	&0.654 	&\textbf{0.876} 	&0.798 \\
\!\!car-roundabout\!\!	     &0.851 	&0.717 	&0.726 	&0.708 	&0.614 	&0.777 	&\textbf{0.904}\\
dance-twirl	     &0.492 	&0.471 	&0.444 	&0.117 	&0.099 	&0.318 	&\textbf{0.640}\\
drift-chicane	     &0.033 	&0.457 	&0.243 	&0.119 	&0.018 	&0.331 	&\textbf{0.466} \\
\!\!horsejump-low\!\!	     &0.601 	&0.607 	&0.663 	&0.498 	&0.291 	&0.551 	&\textbf{0.768} \\
libby	             &\textbf{0.776} 	&0.316 	&0.295 	&0.226 	&0.070 	&0.553 	&0.723 \\
mallard-fly	         &0.606 	&0.541 	&0.536 	&0.557 	&0.200 	&0.436 	&\textbf{0.650}\\
motorbike	         &0.563 	&0.713 	&0.506 	&0.451 	&0.340 	&0.687 	&\textbf{0.749}\\
rhino	             &0.782 	&0.794 	&0.716 	&0.736 	&0.694 	&0.812 	&\textbf{0.893}\\
soapbox	             &\textbf{0.789} 	&0.449 	&0.759 	&0.783 	&0.247 	&0.684 	&0.751 \\
stroller	         &0.767 	&0.597 	&0.656 	&0.464 	&0.369 	&0.662 	&\textbf{0.826}\\
surf	             &0.492 	&0.843 	&\textbf{0.941} 	&0.821 	&0.814 	&0.759 	&0.917\\
swing	             &\textbf{0.784} 	&0.648 	&0.115 	&0.511 	&0.098 	&0.104 	&0.765\\
tennis	             &0.737 	&0.623 	&0.765 	&0.482 	&0.074 	&0.576 	&\textbf{0.826} \\
\hline
\hline
Avg. (entire)         &0.665    &0.631    &0.607    &0.556    &0.358    &0.596    &\textbf{0.736}\\
\hline
\end{tabular}
}
\hfill\mbox{}
\caption{ IoU score on a representative subset of
the DAVIS dataset \cite{Perazzi2016}, and the average computed over all 50 video sequences. The best results are \textbf{boldfaced}.
}\label{table1}
\end{table}

\noindent\textbf{Quantitative Results~}Standard Intersection-over-Union (IoU) metric is employed for quantitative evaluation. Given a segmentation mask $M$ and ground-truth $G$, IoU is computed via $\frac{M\cap G}{M\cup G}$.
We compare the proposed STV against
various state-of-the-art alternatives: BVS \cite{maerki2016}, FCP \cite{Perazzi2015}, JMP \cite{Fan2015}, SEA \cite{Ramakanth2014}, TSP~\cite{chang2013}, HVS \cite{grundmann2010}, JOT \cite{Wen_2015_CVPR}, and OFL \cite{Tsai2016}.

In Table \ref{table1}, we report IoU score on a \textit{representative} subset of the DAVIS dataset.
As shown, the proposed STV performs superior on most video sequences.
And STV achieves the highest average IoU score (\textbf{0.736}) over all the 50 video sequence of the DAVIS dataset, which
demonstrates significant improvement over previous methods. 

We further report quantitative results on Segtrack-V2 \cite{li2013} dataset in Table \ref{table3}. The results consistently demonstrate the favorable performance of the proposed method.

\begin{table}
\centering
\resizebox{0.49\textwidth}{!}{
\begin{tabular}{|c||ccccccc|}
\hline
\!\!Method\!\! &BVS &OFL &SEA  &FCP &HVS &JOT &\textbf{STV} \\
 \hline
  \hline
IoU    &\!\!0.584\!\! &\!\!0.675\!\! &\!\!0.453\!\!  &\!\!0.574\!\! &\!\!0.518\!\! &\!\!0.718\!\! &\!\!\textbf{0.781}\!\!\\
\hline
\end{tabular}
}
\hfill\mbox{}
\caption{Average IoU score for SegtrackV2 dataset. The
best results are \textbf{boldfaced}.}\label{table3}
 \vspace{-2mm}
\end{table}

\begin{figure*}
  \centering
      \includegraphics[width=0.95 \linewidth]{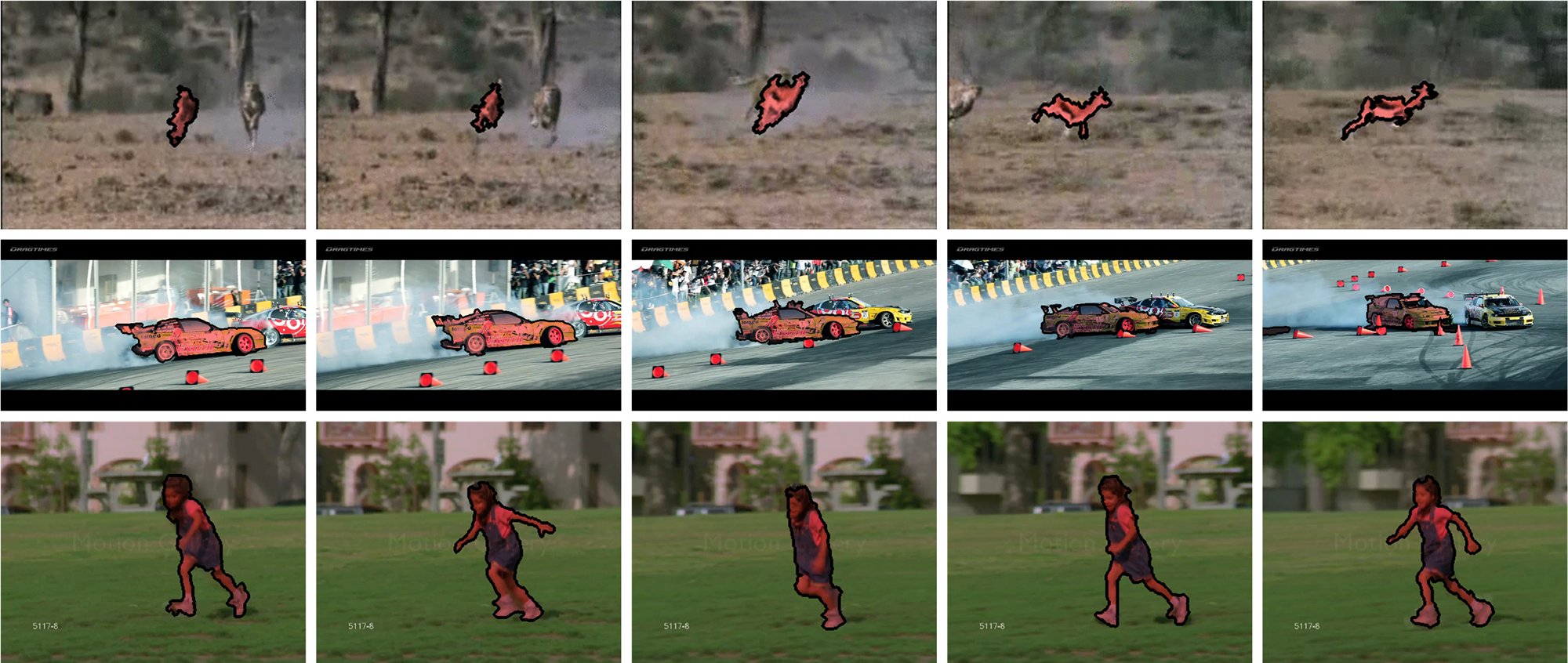}
\caption{Qualitative segmentation results on representative video sequences from SegTrack-V2 \cite{li2013} (from top to bottom: \textit{cheetah1}, \textit{drift1}, and \textit{girl}). The initial masks are presented in the first row. }
 \label{fig:fig11}
 \vspace{-5mm} 
\end{figure*}

\noindent\textbf{Qualitative Results~}Qualitative video segmentation results
for video sequences from the DAVIS dataset \cite{Perazzi2016} and SegTrack-V2 \cite{li2013}
are presented in Fig.~\ref{fig:fig5} and Fig.~\ref{fig:fig11}.
With the first frame as initialization, the proposed algorithm has the ability to segment the objects with fast
motion patterns (\textit{breakdance-flare} and \textit{cheetah1}) or large shape deformation
(\textit{dog-agility}). It also produces accurate segmentation maps even when
the foreground suffers occlusions (\textit{libby}).

\subsection{Validation of the Proposed Algorithm}
\label{sec:VPA}
In this section, we offer more detailed exploration for the proposed approach in several aspects with DAVIS dataset \cite{Perazzi2016}. We test the values of important parameters, verify basic assumptions of the proposed algorithm, evaluate the contributions from each part of our approach, and perform runtime comparison.
\begin{figure}
  \centering
      \includegraphics[width=0.95 \linewidth]{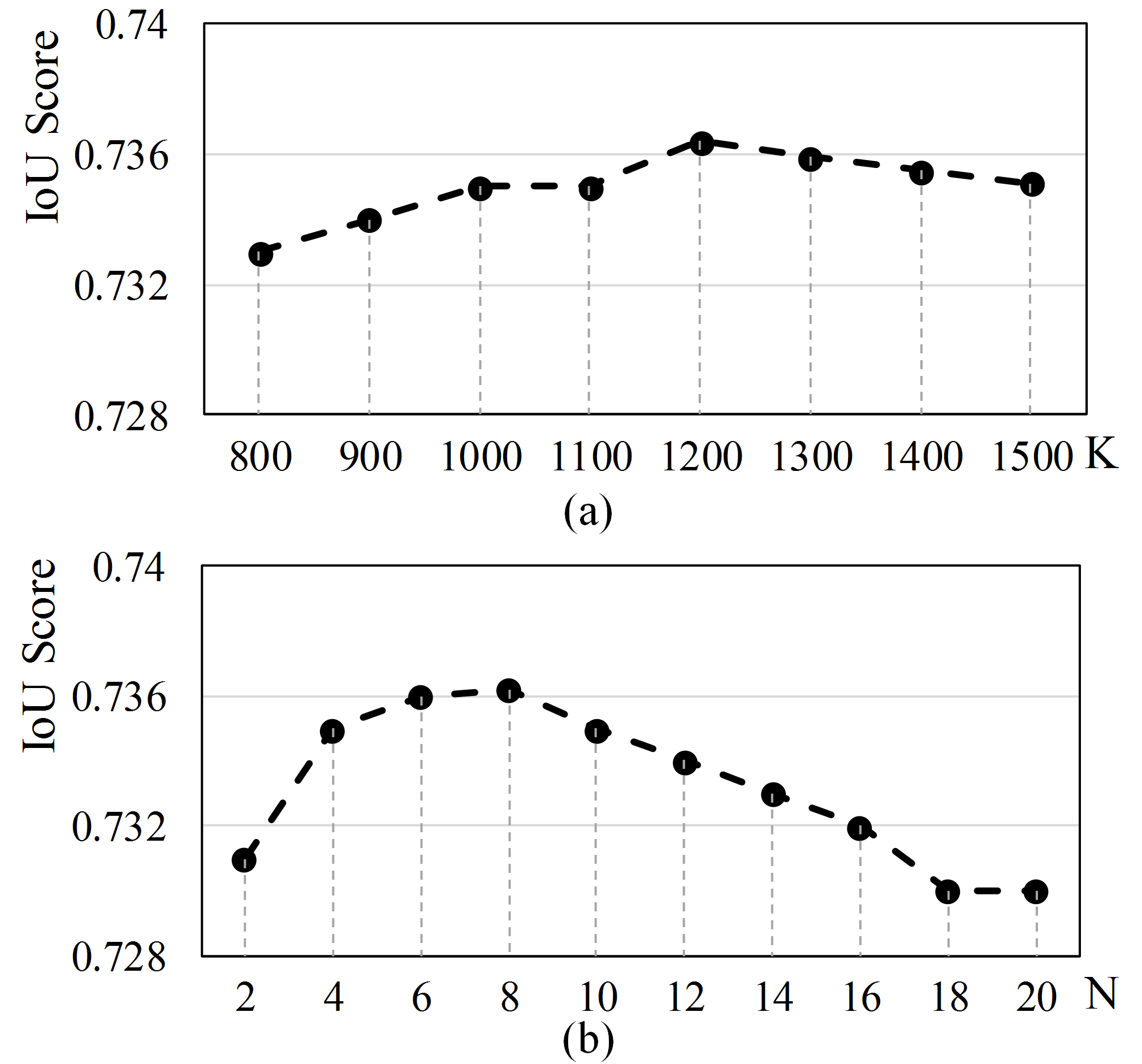}
\caption{Parameter selection for number of spatial grids $K$ (a) and the number of the number of the NNs $N$ (b). The IoU score is plotted as a function of a variety of $K$s ($N$s).}
 \label{fig:fig6} 
\end{figure}

\noindent\textbf{Parameter Verification~}We study the influence of the needed input parameter: number of spatial grids $K$, of our super-trajectory algorithm in Sec.~\ref{sec:supertrajectorygeneration}.
We report the performance by plotting the IoU value  of the segmentation
results as functions of a variety of $Ks$, where we vary $K = \{800,900,{\cdots},1500\}$.
As shown in Fig.~\ref{fig:fig6} (a), the performance increases with finer super-trajectory clustering in spatial domain  ($K\!\!\uparrow$). However, when we further increase $K$, the final performance does not change obviously.
We set $K=1200$ where the maximum performance is obtained. Later, we investigate the influence of parameter $N$, which indicates the number of the NNs of a region in Sec.~\ref{sec:segmentation}. We plot IoU score with varying $N = \{2,4,{\cdots},20\}$ in Fig.~\ref{fig:fig6} (b), and set $N = 8$ for achieving best performance.

\noindent\textbf{Ablation Study~}To quantify the improvement obtained with our proposed trajectories in Sec.~\ref{sec:trajectorygeneration},
we compare to two baseline trajectories: LTM \cite{Fragkiadaki12} and DAD \cite{wang2011} in our experimental results. LTM is widely used for motion segmentation and DAD shows promising performance for action detection. To be fair, we only replace our trajectory generation part with above two methods, estimate optical flow via LDOF~\cite{BroxT2011} and keep all other parameters fixed. From the comparison results in Table.~\ref{table5}. we can find that, compared with classical trajectory methods \cite{Fragkiadaki12,wang2011}, the proposed trajectory generation approach is preferable.

\begin{table}
\centering

\begin{tabular}{|c||ccc|}
\hline
\!\!Method\!\! &LTM &DAD &\textbf{STV}\\
\hline
\hline
IoU    &\!\!0.718\!\! &\!\!0.654\!\! &\!\!\textbf{0.736}\!\!\\
\hline
\end{tabular}

\hfill\mbox{}
\caption{Average IoU score for DAVIS dataset with comparison to two trajectory methods: LTM \cite{Fragkiadaki12} and DAD \cite{wang2011}. The
best results are \textbf{boldfaced}.}\label{table5}
\vspace{-6mm}
\end{table}

\section{Conclusions}
\label{sec:conclusion}
This paper introduced a video segmentation approach by representing video as super-trajectories.
Based on DPC algorithm, compact trajectories are efficiently grouped into super-trajectories.
Occlusion and drift are naturally handled by our trajectory generation method based on a probabilistic model.
We proposed to perform video segmentation on super-trajectory level. Via reverse tracking points and leveraging the property of region re-occurrence, the algorithm is robust for many segmentation challenges.
Experimental results on famous video segmentation datasets \cite{Perazzi2016,li2013} demonstrate that our approach outperforms current state-of-the-art methods.

{\small
\bibliographystyle{ieee}
\bibliography{egbib}
}

\end{document}